\pdfoutput=1
\documentclass{article}
\PassOptionsToPackage{hidelinks}{hyperref}
\usepackage{ijcai25}

\usepackage{times}
\usepackage{soul}
\usepackage{url}
\usepackage[small]{caption}
\usepackage{graphicx}
\usepackage{amssymb}   
\usepackage{amsmath}    
\usepackage{amsfonts}   
\usepackage{mathrsfs}
\usepackage[table]{xcolor} % 加载xcolor包并启用table选项

\usepackage{booktabs}
\usepackage{algorithm}
\usepackage[noend]{algorithmic}
\usepackage{color}
\definecolor{Red}{RGB}{192, 0, 0}
\definecolor{Blue}{RGB}{0, 0, 192}
\definecolor{Green}{RGB}{0, 192, 0}

\usepackage[switch]{lineno}
\usepackage{footmisc}

\title{Leveraging Personalized PageRank and Higher-Order Topological Structures for Heterophily Mitigation in Graph Neural Networks}

\author{
Yumeng Wang$^{1}$
\and
Zengyi Wo$^{1,2}$ \and
Wenjun Wang$^{1,2,3}$ \and
Xingcheng Fu$^{4}$  \and
Minglai Shao$^{1}$ \\
\affiliations
$^1$School of New Media and Communication, Tianjin University, Tianjin, China\\
$^2$College of Intelligence and Computing, Tianjin University, Tianjin, China\\
$^3$Yazhou Bay Innovation Institute, Hainan Tropical Ocean University, Sanya, Hainan, China\\
$^4$Key Lab of Education Blockchain and Intelligent Technology, Guangxi Normal University, China\\
\emails
\{ymwang, wozengyi1999, wjwang, shaoml\}@tju.edu.cn, fuxc@gxnu.edu.cn
}

\begin{document}
\maketitle
\begin{abstract}
Graph Neural Networks (GNNs) excel in node classification tasks but often assume homophily, where connected nodes share similar labels. This assumption does not hold in many real-world heterophilic graphs. Existing models for heterophilic graphs primarily rely on pairwise relationships, overlooking multi-scale information from higher-order structures. This leads to suboptimal performance, particularly under noise from conflicting class information across nodes.
To address these challenges, we propose \textbf{HPGNN}, a novel model integrating \textbf{H}igher-order \textbf{P}ersonalized PageRank with \textbf{G}raph \textbf{N}eural \textbf{N}etworks. HPGNN introduces an efficient high-order approximation of Personalized PageRank (PPR) to capture long-range and multi-scale node interactions. This approach reduces computational complexity and mitigates noise from surrounding information. By embedding higher-order structural information into convolutional networks, HPGNN effectively models key interactions across diverse graph dimensions.
Extensive experiments on benchmark datasets demonstrate HPGNN's effectiveness. The model achieves better performance than five out of seven state-of-the-art methods on heterophilic graphs in downstream tasks while maintaining competitive performance on homophilic graphs. HPGNN's ability to balance multi-scale information and robustness to noise makes it a versatile solution for real-world graph learning challenges. Codes are available at https://github.com/streetcorner/HPGNN.
\end{abstract}

% 提出别人的，没有被解决的缺点。
\section{Introduction}
Graph Representation Learning (GRL) tackles the challenge of capturing complex relational information in non-Euclidean structured data, enabling advanced machine learning applications on graphs~\cite{hamilton2017representation,hamilton2020graph}. Graph Neural Networks (GNNs), a cornerstone of GRL, were originally developed under the homophily assumption, where connected nodes share similar properties. Early GNN models focused on homophilic graphs~\cite{velivckovic2017graph,xu2018powerful}, yet many real-world networks exhibit heterophily, with connections spanning dissimilar nodes. The performance of GNNs often depends on the graph's homophily level~\cite{pei2020geom}, driving recent efforts to address heterophily~\cite{jin2021universal,zhu2022does,wu2023learning,qiu2024refining,wo2024graph}. 

Despite these advances, most prior work centers on pairwise node relationships~\cite{majhi2022dynamics}, which fail to capture the multi-scale, complex interactions prevalent in real-world scenarios while higher-order structures better reflect real-world complex systems~\cite{benson2016higher}. To overcome this limitation, higher-order network methods, such as hypergraphs and simplicial complexes (SCs), have emerged. Hypergraphs generalize graphs by allowing edges to connect multiple nodes but often neglect intra-hyperedge relationships, limiting their robustness to noise compared to SCs~\cite{antelmi2023survey}. Grounded in algebraic topology~\cite{spanier1989algebraic}, SCs model multidimensional relationships, offering richer node context~\cite{hajij2021simplicial} and proving effective in domains like propagation~\cite{iacopini2019simplicial}, sensor networks~\cite{schaub2021signal}, and social systems~\cite{fu2024higher,benson2016higher}.

A central challenge in higher-order neural networks lies in constructing node relationships using SCs. Recent approaches leverage the Hodge Laplacian from SC theory~\cite{ebli2020simplicial,roddenberry2021principled,yang2022simplicial}, but its computation can reach \( \mathcal{O}(n^3) \) complexity in dense graphs, posing scalability issues. Meanwhile, \textbf{Personalized PageRank (PPR)}, a random walk variant with teleportation~\cite{page1999pagerank}, efficiently captures local node information. Recent studies integrate PPR with GNNs to enhance representations while preserving efficiency~\cite{gasteiger2018predict,chien2020adaptive,jack2023s}, though they often overlook multi-level, higher-order interactions.

In this paper, we propose \textbf{HPGNN}, a novel framework that combines higher-order topological structures with Personalized PageRank to advance graph representation learning, particularly for heterophilic graphs. Unlike methods limited to node-to-node interactions, HPGNN incorporates relationships across multiple simplicial complexes. Our approach comprises two key components: 1) \textbf{Higher-order Personalized PageRank (HiPPR)}, which efficiently computes long-range interactions while reducing noise, and 2) \textbf{Higher-order Adaptive Spectral Convolution (HiASC)}, which extends spectral convolution to capture higher-dimensional relationships via SCs.
Our contributions are threefold:
\begin{itemize}
    \item HPGNN pioneers the integration of higher-order information into PPR approximation, bridging simplicial complex theory with personalized random walks.
    
    \item We introduce an adaptive PPR matrix operator that encodes interactions among higher-order graph structures.
    \item Extensive experiments on real-world datasets demonstrate that HPGNN achieves better performance than five out of seven state-of-the-art models.
\end{itemize}
\section{Related Work}

In this section, we provide a brief discussion of related work on learning from graphs with heterophily and personalized pageRank.

% \subsection{Personalised Pagerank}  Personalized PageRank (PPR) is a key problem in network science, with recent advancements focusing on efficient approximation to improve scalability and precision~\cite{yang2024efficient}. One notable approach, APPNP, integrates PPR with Graph Neural Networks (GNNs) through a prediction-then-propagate strategy, improving node representation learning by efficiently approximating PPR~\cite{gasteiger2018predict}. Additionally, GPRGNN introduces an adaptive mechanism that dynamically adjusts teleportation probabilities, allowing for better capture of personalized node features across diverse graph structures~\cite{chien2020adaptive}. Yet, both methods fail to effectively mitigate surrounding noise in heterophily graphs.
\subsection{GNNs with Heterophily}
Recent GNNs for heterophilous graphs break traditional homophily assumptions, APPNP~\cite{gasteiger2018predict} decouples feature transformation and propagation, while GPRGNN~\cite{chien2020adaptive} adaptively combines graph filters for task-specific aggregation. They enable long-range information propagation and reduce over-smoothing, preserving node-specific features. BernNet expands by leveraging Bernstein polynomial filters to effectively balance low and high frequency signals to handle label inconsistency in heterophily~\cite{chen2022bscnets}. Most recently, POLYGCL~\cite{chen2024polygcl} improves graph contrastive learning via structure-aware sampling and polynomial propagation, preserving structural roles and handling heterophily. 
Nonetheless, these methods generally lack a high-order perspective to stably suppress boundary noise caused by conflicting class in-
formation across nodes. 

\subsection{Higher-order Simplicial Neural Networks} 
Simplicial Neural Networks focus on efficiently modeling higher-order structures and embedding them into neural networks. BScNets use block operators to capture dependencies across multi-order structures.~\cite{chen2022bscnets}. Another approach extends the block matrix concept, defining an inter-order adjacency matrix for transformations between simplices~\cite{zhou2024facilitating}. SCN introduces random walks on simplices, enabling comprehensive simplex learning~\cite{wu2023simplicial}. HiGCN extends SCN with two-step random walks, capturing dis                                                                                                                                                       tant higher-order relationships~\cite{huang2024higher}. Despite these advancements, traditional random walk methods face issues like rapid information diffusion and lack of personalized control, hindering effective node importance differentiation. 

\section{Preliminary}

This subsection provides the foundational concepts and problem setting for our work. 

\subsection{Background}
Our work builds on foundational concepts of graphs, graph heterophily, simplicial complexes, and clique complex lifting.

\paragraph{Graph.}Consider an undirected, unweighted graph $G = (V, E)$, where $V = \{v_1, v_2, \dots, v_N\}$ is a finite set of $N = |V|$ nodes, and $E \subseteq V \times V$ is the set of $M = |E|$ edges. Let $X \in \mathbb{R}^{N \times F}$ denote the node feature matrix, where $F$ is the feature dimension, and $A \in \{0, 1\}^{N \times N}$ represent the adjacency matrix, with $A_{ij} = 1$ if $(v_i, v_j) \in E$ and $0$ otherwise.

\paragraph{Graph Heterophily.} Graph homophily reflects the tendency of connected nodes to share similar attributes, while heterophily indicates connections between dissimilar nodes~\cite{mcpherson2001birds}. We quantify homophily using the node homophily ratio:
\begin{equation}
H_{\text{node}} = \frac{1}{N} \sum_{v \in V} \frac{\left|\{(u, v) \in E : y_u = y_v\}\right|}{d_v},
\label{eq:node_homophily}
\end{equation}
where $y_v$ is the label of node $v$, and $d_v$ is its degree. The ratio $H_{\text{node}} \in [0, 1]$ indicates the level of homophily, with lower values signifying stronger heterophily. This metric is critical for understanding network properties in heterophilic settings~\cite{pei2020geom}.

\paragraph{Simplicial Complexes.} A simplicial complex $\mathcal{K}$ is a finite collection of simplices satisfying: 1) a $k$-simplex $\sigma^k = \{v_0, v_1, \dots, v_k\}$ is the convex hull of $k+1$ vertices; 2) if $\sigma^k \in \mathcal{K}$, all its subsets (faces) are in $\mathcal{K}$; and 3) $\mathcal{K}$ is finite~\cite{battiston2020networks}. Simplicial complexes model higher-order interactions, where a 0-simplex is a vertex, a 1-simplex an edge, a 2-simplex a triangle, and so forth, enabling the study of topological relationships.

\subsection{Problem Setting}
In semi-supervised node representation learning, the goal is to learn an encoder $f(\cdot): \mathbb{R}^{N \times F} \times \mathbb{R}^{N \times N} \to \mathbb{R}^{N \times D}$ that maps the node feature matrix $X$ and adjacency matrix $A$ to a representation matrix $Z = f(X, A) = \{z_1, z_2, \dots, z_N\}$, where $z_i \in \mathbb{R}^D$ is the $D$-dimensional representation of node $v_i$. These representations are optimized for downstream tasks, such as node classification or clustering, particularly in graphs with varying levels of homophily or heterophily.

\begin{figure*}[ht]
 \centering
  \includegraphics[width=6.4 in]{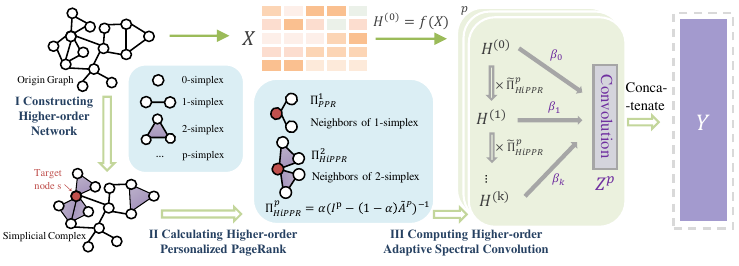}
    \caption{\textsc{HPGNN Architecture.} A modular framework comprising: (1) clique complex lifting to construct higher-order topological structures, (2) Higher-order Personalized PageRank (HiPPR) for robust long-range dependency modeling and (3) Higher-order Adaptive Spectral Convolution (HiASC) to capture multi-dimensional interactions.}
    \label{fig.model}
  
\end{figure*}

\section{Methodology}
\paragraph{Overview.} We propose the \textbf{H}igher-order \textbf{P}ersonalized PageRank \textbf{G}raph \textbf{N}eural \textbf{N}etwork (HPGNN) to address heterophily and model complex node interactions in graph representation learning. As shown in Figure~\ref{fig.model}, HPGNN integrates three components: (1) Clique complex lifting~\cite{schaub2020random} to transform the original pairwise graphs into simplicial complexes, providing more comprehensive views of the graph structure; (2) Higher-order Personalized PageRank (HiPPR) to model long-range dependencies while mitigating noise in heterophilic graphs; and (3) Higher-order Adaptive Spectral Convolution (HiASC) to model higher-dimensional relationships via spectral methods. By combining HiPPR and HiASC, HPGNN produces robust node representations that excel in heterophilic settings and remain competitive in homophilic ones, enabling superior performance in downstream tasks.

\subsection{Constructing Higher-order Network}
To capture higher-order interactions, we transform a graph $G = (V, E)$ into simplicial complexes(SCs) via clique complex lifting~\cite{schaub2020random}. Let $C(G)$ denote the set of all cliques in $G$, where a clique $C_i \subseteq V$ is a complete subgraph. Each $p$-clique in $C(G)$ is mapped to a $(p-1)$-simplex, where 2-simplices are the 3-cliques in the graph, forming the SC $\mathcal{K} = \bigcup_{C_i \in C(G)} \sigma_i$. This process encodes multi-node interactions beyond pairwise edges.

\subsection{Higher-order Personalized PageRank}
PPR is a variant based on random walks, it modified the original PageRank algorithm to incorporate a ``teleport''mechanism that allows us to personalize the random walk starting from a specific root node~\cite{page1999pagerank}. This adaptation ensures the influence of the root node is reflected, preserving the local structure around it even in the long run.

\begin{algorithm}[tb]
\caption{HiPwrPushSOR: Higher-order Personalized PageRank Approximation}
\label{alg:HiPwrPushSOR}
\textbf{Input:} Simplicial complex $\mathcal{K}$, source node $s \in \mathcal{K}$, number of nodes $n$, number of simplices $m$ \\
\textbf{Parameter:} Precision $\epsilon$, teleport probability $\alpha \in (0, 1)$, error threshold $\lambda > 0$, relaxation parameter $\omega > 0$
\\
\textbf{Output:} Estimated PPR vector $\hat{\pi}_s \in \mathbb{R}^n$

\begin{algorithmic}[1]
\STATE Initialize: $\text{epochNum} \gets 8$, $\text{scanThreshold} \gets \lfloor n/4 \rfloor$

\STATE Set $\hat{\pi}(s, \sigma) \gets 0$, $r(s, \sigma) \gets 0$ for all $\sigma \in \mathcal{K}$; $r(s, s) \gets 1$
\STATE Initialize FIFO queue $Q \gets \emptyset$; $Q.\text{append}(s)$
\STATE Set $r_{\text{max}} \gets \lambda / m$; $r_{\text{sum}}=1$
\WHILE{$Q \neq \emptyset$ and $|Q| \leq \text{scanThreshold}$ and $r_{\text{sum}} > \lambda$}
    \STATE $\sigma \gets Q.\text{pop}()$
    \STATE $\hat{\pi}(s, \sigma) \gets \hat{\pi}(s, \sigma) + \omega \alpha r(s, \sigma)$
    \STATE $r_{sum} \gets r_{sum} - \omega \alpha r(s, \sigma)$
    \FOR{each $\tau \in N_{\text{out}}(\sigma)$}
        \STATE $r(s, \tau) \gets r(s, \tau) + \frac{\omega (1 - \alpha) r(s, \sigma)}{d_{\sigma}}$
        \IF{$\tau \notin Q$ } \STATE $Q.\text{append}(\tau)$
        \ENDIF
    \ENDFOR
\ENDWHILE
\IF{$r_{\text{sum}} > \lambda$}
    \STATE \textit{/* Switch to sequential scan */}
    \FOR{$i = 1$ to $\text{epochNum}$}
        \STATE $r'_{\text{max}} \gets \lambda^{i / \text{epochNum}} / m$
        \WHILE{$r_{sum} > m \cdot r'_{\text{max}}$}
            \FOR{each $\sigma \in \mathcal{K}$}
                \IF{$|r(s, \sigma)| \geq \epsilon d_{\sigma}$}
                    \STATE $\hat{\pi}(s, \sigma) \gets \hat{\pi}(s, \sigma) + \omega \alpha r(s, \sigma)$
                    \FOR{each $\tau \in N_{\text{out}}(\sigma)$}
                        \STATE $r(s, \tau) \gets r(s, \tau) + \frac{\omega (1 - \alpha) r(s, \sigma)}{d_{\sigma}}$
                    \ENDFOR
                \ENDIF
            \ENDFOR
        \ENDWHILE
    \ENDFOR
\ENDIF
\RETURN $\hat{\pi}_s = [\hat{\pi}(s, \sigma)]_{\sigma \in \mathcal{K}}$
\end{algorithmic}
\end{algorithm}

On pairwise graphs, The root node $s$ is defined via the teleport vector $i_s$, which is a one-hot indicator vector. The PPR of node $s$ can be computed as: $\pi_{\text{ppr}}(i_s) = \alpha i_s + (1 - \alpha) \tilde{A} \pi_{\text{ppr}}(i_s)$, with the teleport (or restart) probability $\alpha \in (0,1]$. By solving the equation and substituting the indicator vector $i_s$ with the unit matrix $I_s$, we obtain the PPR matrix:

\begin{equation}
\Pi_{\text{PPR}} = \alpha \left( I_n - (1 - \alpha) \tilde{A} \right)^{-1},
\end{equation}
where $\tilde{A}$ is the normalized adjacency matrix, defined as $\tilde{A}= AD^{-1}$, $A$ is the adjacency matrix of the graph and $D$ is the degree matrix. $I_n$ is the identity matrix of size $n\times n$, and $n$ is the number of nodes in the graph. The matrix $\Pi_{\text{ppr}}$ thus represents the personalized influence of node $s$ on all other nodes in the graph.

\subsubsection{Higher-order Information Integration in PPR}
After incorporating the higher-order information, we design to treat the node as a simplice to obtain higher-order PPR. The HiPPR vector $\Pi_{\text{ppr}}$ for a given simplix $\sigma$ can be derived from the concept of a random walk with restarts. Specifically, for the $p$-order simplicial complex, the corresponding Higher-order Personalized PageRank (HiPPR) Matrix can be calculated as: 
\begin{equation} 
\Pi^{p}_{\text{HiPPR}} = \alpha \left( I^p - (1 - \alpha) \tilde{A}^p \right)^{-1}, 
\end{equation} 
where $I^p$ is the identity matrix corresponding to the number of simplices of order $p$ and $\tilde{A}_p$ is the normalized adjacency matrix incorporating the higher-order simplicial structure, representing the adjacency matrix between $0$-simplices(nodes) and $p$-simplices (where $p>0$) . This innovative formula extends the basic idea of PPR to account for higher-order relationships within the simplicial complex.

In practical computations, approximate algorithms balance speed and accuracy. Recent key approximate approaches~\cite{yang2024efficient} include Monte Carlo, Power Iteration, and Forward Push. We chose to improve the PwrPushSOR method~\cite{chen2023accelerating}, which combines Power Iteration and Forward Push, achieving better error and runtime on undirected graphs. It computes PPR vectors for each node $v$ with $O(n \cdot d(v)_{avg})$ time and $O(n)$ space complexity. The innovative algorithm HiPwrPushSOR used to approximate the higher-order PPR vector is presented in Algorithm 1. The input is a single root node $s$, and the output is the PPR vector we aim to compute. For SCs with order $p$, we compute $p$ PPR vctor for the root node.

The input includes the SC $\mathcal{K}$, the source node $s$ and parameters $\alpha,  \epsilon, \lambda$ and $\omega$. Where $\alpha$ controls the transition probability, which affects the range of information propagation in the PPR algorithm; $\epsilon$ determines the node activation criteria for queue insertion; $\lambda$ regulates the stopping condition to guarantee computational precision; and $\omega\in (0,2)$ is the Successive Over-Relaxation(SOR) parameter to control the intensity of the update at each iteration. In this study, we adopt the value recommended for undirected graphs~\cite{chen2023accelerating}.

In the new algorithm, we iterate all nodes (0 to n-1), treating each as the source node $s$ to calculate its PPR vector, which reflects its relationships with all other nodes. By computing PPR vectors for every node, we obtain the PPR matrix. After computing the PPR matrix for each order $p$, $\Pi^{p}_{\text{HiPPR}}$is obtained. Moreover, excessive propagation of higher-order information in node classification can cause redundancy and increased computational cost. To address this, we introduce a constraint in HiPPR, modeling nodes propagating to higher-order neighbors and back, thus efficiently controlling higher-order information flow.

\subsection{Higher-order Adaptive Spectral Convolution}

\subsubsection{Spectral graph convolution} 

Served as the foundation for many spectral GNNs~\cite{bianchi2021graph}, spectral graph convolution is typically unified under the expression $Z = g(L)\cdot X$, where $Z$ represents the output feature matrix and $g(L)$
is the graph filter applied to the graph Laplacian matrix $L$. The core of spectral graph convolution lies in the design of the graph filter $g$, which governs how graph signals are manipulated in the spectral domain. 

For a given graph Laplacian matrix $L = U \Lambda U^{\top}$, the eigenvalue matrix $\Lambda = \text{diag}(\lambda_1, \dots, \lambda_m)$ captures the frequency components of the graph signal, while $U = [u_1, u_2, \dots, u_m]$ represents the eigenvectors corresponding to these frequencies. The graph Fourier transform for a graph signal $x$ is given by $\hat{x} = U^{\top} x$. The goal of spectral graph convolution is to manipulate the frequency response of the graph signal by designing an appropriate graph filter $g(L)$. The graph convolution in the spectral domain can then be expressed as:
\begin{equation}
z = g(L) x  = U  g(\Lambda)U^{\top} x. 
\end{equation}

However, computing the spectral decomposition of large-scale networks is computationally prohibitive. To address this challenge, a practical approach is to approximate the graph filter using polynomial expansions, which avoids direct spectral decomposition. Specifically, the graph filter $g(\lambda)$ is often approximated as a truncated polynomial: $g(\lambda) = \sum_{k=0}^{K} \alpha_k \lambda^k$, where $\alpha_k$ are learnable coefficients and 
$K$ denotes the highest polynomial order considered. This polynomial approximation allows for efficient spectral convolution, bypassing expensive eigen-decompositions.
% \vspace{1em}  % 插入行空白

\subsubsection{Incorporating Higher-order Structures}

Expanding upon the standard spectral graph convolution, we introduce higher-order spectral convolutions by integrating SCs. In this context, the graph laplacian is generalized to higher-order laplacians, which model interactions within simplices—structures that generalize edges to higher-dimensional relationships like triangles and tetrahedra. For each order of the SCs, we compute a PPR laplacian matrix, which captures the connectivity relationships at different orders. The convolution operation for each Higher-order Laplacian is defined as:

\begin{equation}
\quad g_p(L_p) = \sum_{k=0}^{K} \beta_{p,k} L_p^k,
\end{equation}
where $g_p(L_p)$ represents the graph filter applied to the $p$-th order Laplacian,  $p$ corresponds to the order of the SCs. $\beta_{p,k}$ are learnable coefficients that capture the influence of neighbors at different hops in the higher-order SCs, and $k$ determines the polynomial degree for each PPR laplacian matrix.

The convolution across higher-order graphs allows us to encode richer information from neighbors at varying distances within the graph. For a signal $X \in \mathbb{R}^{n \times d} $ with $d$ characteristics per node, we first extracts hidden state features for each node and then uses HiPPR to propagate them:
 
\begin{equation}
 H^{0}=f(X), H^{k}= \tilde{\Pi}^p_{HiPPR} H^{(k-1)},
\end{equation}

\begin{equation}
Z^p =\sum_{k=0}^{K} \beta_{p,k}H^{p,k}, Y = \rho(\big\|_{p=1}^{P} Z^p),
\end{equation}
where $Z^p$ represents the output combining contributions from the $k$-th hop, $\beta_{k}$ are $k$ learnable coefficients, $\tilde{\Pi}^p_{HiPPR}$  captures neighborhood information at order $p$, $\tilde{\Pi}^p_{HiPPR}$ is the normalised version of ${\Pi}^p_{HiPPR}$, formally defined as $\tilde{\Pi}^p_{HiPPR}=D^{-1/2}{\Pi}^p_{HiPPR} D^{-1/2}$, where $D$ is the diagonal degree matrix. $X$ is the input feature matrix, $\rho$ is the simplified linear functions suggested by\cite{wang2022powerful}, and $\big\|$ denotes concatenation.
% \vspace{1em}  % 插入行空白
\subsubsection{Higher-order Spectral Convolution Formulation}
To implement the process, we refer to the work of~\cite{wang2022powerful}, who simplified the functions $\rho$ into linear forms. Building on this, we integrate our novel design of the PPR operator into Higher-order spectral convolutions, leading to the following formulation for the model output:

\begin{equation}
Y = \big\|_{p=1}^{P} \big(\sum_{k=0}^{K} \beta_{p,k} \tilde{\Pi}^p_{HiPPR} X \Theta_p \big) W,
\end{equation}
where $\beta_{p,k}$, $\Theta_p$ and $W$ are learnable parameters,  $\big\|$ denotes concatenation, and $P$ refers to the highest order of the simplicial complex under consideration. 

By incorporating higher-order interactions through our proposed PPR-based filter, this framework extends the capabilities of spectral graph convolution, enabling the model to capture complex dependencies and multiscale information from the graph structure. This adaptive spectral convolution method not only enhances the representation of graph signals but also improves the scalability and efficiency of the model.

\section{Experiments}
We first give the experimental setup, and then compare the performance of our method with baselines on node classification. After that, we analyse the performance of different components of HPGNN and give the parameter analysis. Finally, we provide a visualization task with t-SNE.
\subsection{Experimental Setup}

\subsubsection{Datasets}
We adopt seven real world datasets to evaluate the performance of our proposed HPGNN, including two categories:

(1) Homophilic Graphs: Citation graphs Cora and Citeseer~\cite{sen2008collective}; Social network Photo~\cite{mcauley2015image}.

\begin{table}[b]
\centering
\fontsize{10}{12}\selectfont  % 设置字体大小为10，行间距为12
\renewcommand{\arraystretch}{1}  % 调整行间距
\setlength{\tabcolsep}{1pt}    % 调整列间距以适应新的布局
\begin{tabular}{l|ccccccc}
\toprule
Datasets & \#Node & \#Edge & \#Feature & \#Class & \#Triang. & \#Homo. \\
\midrule
Cora & 2708 & 10556 & 1433 & 6 & 1630 & 0.810 \\
Citeeseer & 3327 & 9104 & 3703 & 7 & 1167 & 0.736 \\
Photo & 7650 & 119081 & 745 & 8 & 717400 & 0.827\\
\midrule
Cornell & 183 & 298 & 1703 & 5 & 59 & 0.127 \\
Actor & 7600 & 29926 & 931 & 5 & 7121 & 0.219 \\
Texas & 183 & 309 & 1703 & 5 & 67 & 0.087 \\
Wisconsin & 251 & 499 & 1703 & 5 & 118 & 0.192 \\
\bottomrule
\end{tabular}
\caption{Benchmark datasets properties and statistics.}
\label{tab:datasets}
\end{table}

(2) Heterophilic Graphs: Cornell, Texas, and Wisconsin webpage graphs~\cite{pei2020geom}; Actor collaboration network Actor~\cite{pei2020geom}.

We tabulated the basic statistics of the datasets in Table \ref{tab:datasets}. The number of triangles, represents the count of 2-simplex in the graph, where three nodes are mutually connected.

\begin{table*}[htbp] % 使用table*环境
\centering

\fontsize{10}{12}\selectfont  % 设置字体大小为10，行间距为12
\small % Reduce font size
% \begin{tabular}{@{}c | p{1.5cm} p{2cm} p{2cm} |p{2cm} p{2cm} p{2cm} p{2cm}@{}}
\begin{tabular}{@{}c | ccc|cccc @{}}
\toprule
% Method & Cora & Citeseer & Photo & Cornell & Actor & Texas & Wisconsin \\
\textbf{Method} & \textbf{Cora} & \textbf{Citeseer} & \textbf{Photo} & \textbf{Cornell} & \textbf{Actor} & \textbf{Texas} & \textbf{Wisconsin} \\

\midrule
ChebNet & $87.04 \pm 0.10$ & $76.53 \pm 0.10$ & $90.75 \pm 0.14$ & $83.40 \pm 0.47$ & $39.10 \pm 0.87$ & $85.57 \pm 0.36$ & $92.0 \pm 0.30$ \\
GCN & $88.29 \pm 0.14$ & $78.8 \pm 0.12$ & $81.92 \pm 0.25$ & $63.62 \pm 0.81$ & $34.2 \pm 0.10$ & $80.49 \pm 0.54$ & $70.38 \pm 0.62$ \\
GAT & $88.10 \pm 0.13$ & $78.44 \pm 0.13$ & $74.61 \pm 0.90$ & $68.72 \pm 0.37$ & $30.50 \pm 0.45$ & $80.33 \pm 0.32$ & $71.38 \pm 0.40$ \\

\midrule
APPNP & $87.27 \pm 0.17$ & $78.42 \pm 0.19$ & $80.44 \pm 0.28$ & $77.66 \pm 0.51$ & $37.46 \pm 0.93$ & $82.25 \pm 0.44$ & $82.38 \pm 0.39$ \\
GPRGNN & $88.88 \pm 0.15$ & $78.43 \pm 0.19$ & $92.36 \pm 0.11$ & $\underline{87.79 \pm 0.43}$ & $38.75 \pm 0.74$ & $\textbf{93.11} \pm \textbf{0.26}$ & $94.13 \pm 0.32$ \\
BernNet & $87.95 \pm 0.16$ & $78.58 \pm 0.15$ & $93.54 \pm 0.04$ & $85.53 \pm 0.39$ & $38.11 \pm 0.10$ & $89.83 \pm 0.38$ & $\underline{94.50 \pm 0.28}$ \\
HiGCN & $\underline{89.24 \pm 0.91}$ & $78.70 \pm 0.78$ & $\underline{94.13 \pm 0.88}$ & $84.25 \pm 0.44$ & $39.41 \pm 0.12$ & $89.18 \pm 0.46$ & $93.13 \pm 0.26$ \\
POLYGCL & $87.57 \pm 0.62$ & $\underline{79.81 \pm 0.85}$ & $93.12 \pm 0.23$ & $82.62 \pm 0.31$ & $\textbf{41.15} \pm \textbf{0.88}$ & $88.03 \pm 1.80$ & $85.50 \pm 1.88$ \\

\midrule
S2V & $80.15 \pm 0.88$ & $78.21 \pm 0.34$ & $84.33 \pm 0.19$ & $77.63 \pm 0.21$ & $39.22 \pm 0.50$ & $82.12 \pm 0.23$ & $83.48 \pm 0.89$ \\
SNN & $87.13 \pm 1.02$ & $79.87 \pm 0.68$ & $88.27 \pm 0.74$ & $ 70.57  \pm 0.78 $ & $30.59 \pm 0.23$ & $75.16 \pm 0.96$ & $61.93 \pm 0.83$ \\
SGAT & $77.49 \pm 0.79$ & $78.93 \pm 0.63$ & N/A & $ 85.46  \pm 0.36 $ & $36.71 \pm 0.49$ & $89.83 \pm 0.66$ & $81.47 \pm 0.64$ \\
SGAT-EF & $78.12 \pm 0.85$ & $79.16 \pm 0.72$ & N/A & $ 85.21  \pm 0.62 $ & $37.33 \pm 0.58$ & $89.67 \pm 0.74$ & $81.59 \pm 0.81$ \\

\midrule

 \textbf{HPGNN(Ours)} & $\textbf{89.96} \pm \textbf{0.13}$ & $\textbf{79.89} \pm \textbf{0.12}$ & $\textbf{95.19} \pm \textbf{0.14}$ & $\textbf{88.30} \pm \textbf{0.42}$ & $\underline{40.89 \pm 0.10}$ & $\underline{91.48 \pm 0.27}$ & $\textbf{94.75} \pm \textbf{0.33}$ \\

\bottomrule
\end{tabular}

\caption{Node classification results on empirical benchmark networks: mean accuracy (\%) $\pm$ 95\% confidence interval. The best results are in \textbf{bold}, while the second-best ones are \underline{underlined}.}
\label{tab:node-classification}
\vspace{-10pt}
\end{table*}

\subsubsection{Baselines}
We compare HPGNN with three categories of baselines:

(1) Homophily-aware Models: Classical graph convolutional networks based on homophily assumption: ChebNet~\cite{defferrard2016convolutional}, GCN~\cite{xu2018powerful}, GAT~\cite{velivckovic2017graph}. 

(2) Heterophily-aware Models: APPNP~\cite{gasteiger2018predict} mitigates the limitations of local aggregation by employing a fixed personalized pagerank-based propagation scheme, GPRGNN~\cite{chien2020adaptive} generalizes by introducing learnable propagation weights, enabling the model to flexibly capture multi-hop dependencies through adaptive PageRank coefficients, BernNet~\cite{he2021bernnet} adjusts node influences using a probabilistic model to handle heterophily, HiGCN~\cite{huang2024higher} uses flower-petals Laplacian operators to mitigate heterophily challenges, and PolyGCL~\cite{chen2024polygcl}integrates polynomial filters into GCL to effectively capture both local and global structural patterns, leading to improved performance on graphs with varying degrees of homophily. 

(3) Higher-order Models: S2V~\cite{billings2019simplex2vec} and SNN~\cite{ebli2020simplicial} embeds simplicial complexes to capture higher-order relationships, SGAT and SGAT-EF~\cite{lee2022sgat} adapts aggregation weights using attention mechanisms for heterophilic graphs.

\subsubsection{Implementation Details}
We implement our proposed HPGNN with Pytorch framework on four NVIDIA RTX 3090 GPUs. Following~\cite{he2021bernnet}, For all datasets, we randomly split nodes into $60\%$, $20\%$, and $20\%$ for training, validation, and testing, all methods share the same 10 random splits. We train the proposed HiGCN model with the learning rate $lr \in \{0.01, 0.05, 0.1, 0.2, 0.3\}$ and the weight decay $wd \in \{0.0, 0.0001, 0.001, 0.005, 0.1\}$. For all methods, we run 100 times with the same partition and report the average and standard deviation of accuracy values. 

\subsection{Main Experiment Results}
Tables \ref{tab:node-classification} shows the comparison of our proposed HPGNN with SOTAs for node classification task on seven real-world network datasets. It can be concluded that: (1) On heterophilic graphs Cornell, Actor, Texas and Wisconsin, HPGNN shows substantial improvement, particularly in datasets with more complex relationships. This suggests its effectiveness in capturing multiscale node dependencies, which is crucial for heterophilic graph structures. (2) HPGNN maintaining competitive performance on homophilic graphs Cora, Citeseer and Photo, showcasing its versatility and robustness across different types of graph structures, a point further supported by the average standard deviation in the table. (3) Compared to higher-order simplicial complex neural networks, our proposed algorithm shows an overwhelming advantage, mainly because it combines multi-scale features and  adaptive weights to avoid over-smoothing. (4) Though HPGNN performs competitively on Actor and Texas, exceeding the third-best method by at least 1.49 and 1.65 percentage points, respectively. (5) Notably, the top two results across dataset, aside from ours, come from four different algorithms, demonstrating the stability and consistency of our method, highlighting the impact of higher order techniques in achieving this stability. 

\subsection{Ablation Study}
\begin{figure}[t]
 \centering
  \includegraphics[width=3.2 in]{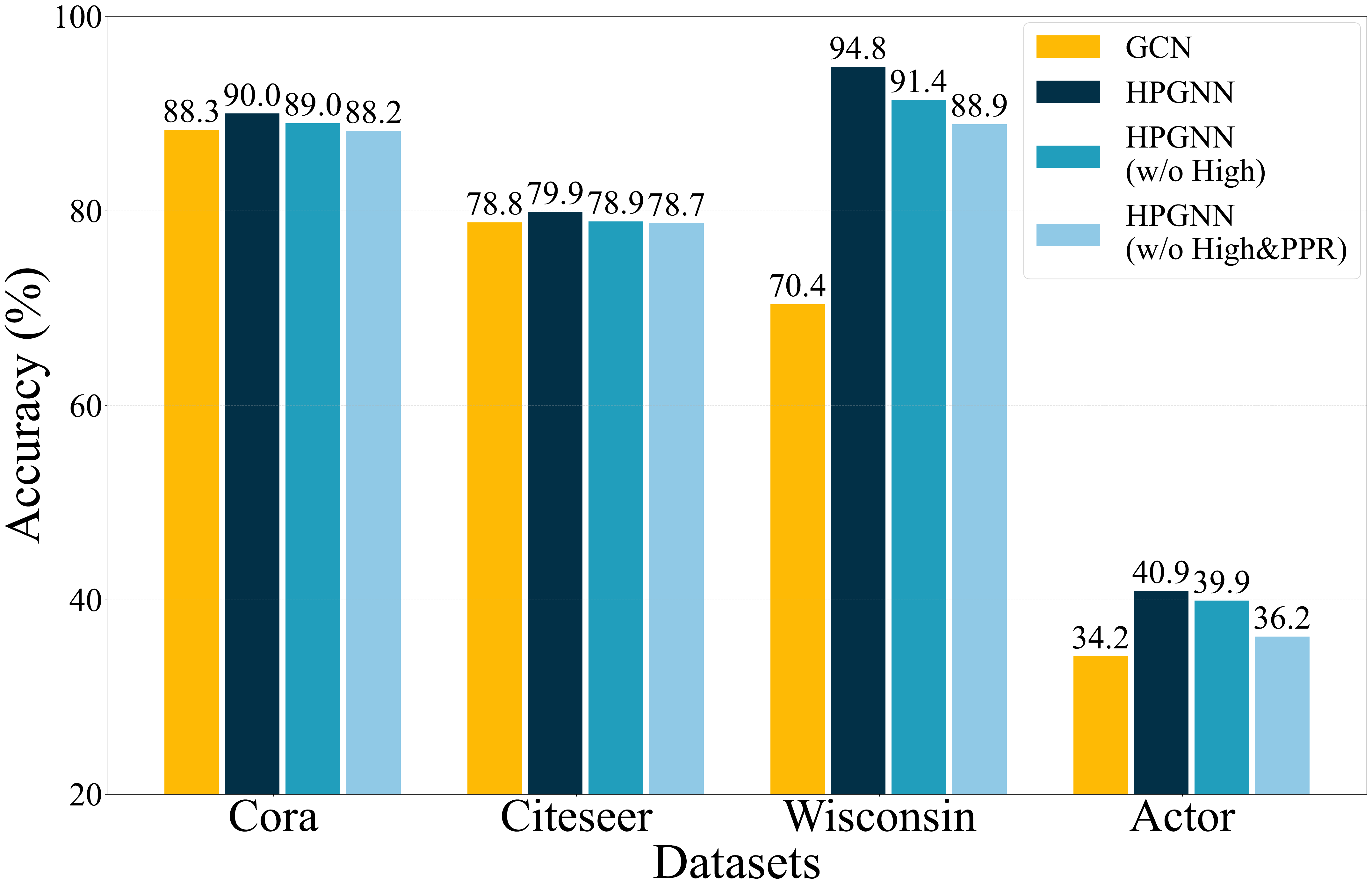}
    \caption{The impact of Higher-order and PPR Modules}
    \label{fig.ablation}
\end{figure}

As shown in Figure ~\ref{fig.ablation}, ablation studies were performed on four datasets: two homophilic graphs (Cora and Citeseer) and two heterophilic graphs (Actor and Wisconsin). To validate the effectiveness of each component in our proposed model, we conducteded several ablation experiments comparing our model with different variants: 1) HPGNN(\textbf{w/o} High): This variant excludes the higher-order structure, using only pairwise PPR weights and convolution for signal filtering between nodes. For this variant, we set the model hyperparameter order to 1 in different datasets.  It shows that removing higher-order structures degrades performance, especially on Wisconsin and Actor, highlighting their role in complex heterophilic graph interactions. 2) HPGNN(\textbf{w/o} High and PPR): This variant removes the PPR mechanism from the pairwise graph, replacing the PPR matrix $\Pi$ with $\Tilde{D}^{-1/2}\Tilde{A}\Tilde{D}^{-1/2}$ as suggested in~\cite{wu2019simplifying}. It can be analyzed that excluding PPR causes a notable drop in performance, particularly on Wisconsin and Actor, underlining its importance in noise filtering and long-range dependency. 

In summary, HPGNN outperforms both variants, demonstrating the effectiveness of the proposed modules. In heterophilic graphs, performance degradation is more significant when higher-order and PPR modules are removed, emphasizing their importance in such environments. Combining both higher-order structures and PPR yields the best results, demonstrating their complementary roles in improving performance.

\subsection{Parameter Analysis}

We estimate the sensitivity of three important hyperparameters, that is, the order $p$, the teleportation parameter $\alpha$ and the convolution layer $k$.
\renewcommand{\arraystretch}{0.85} % 默认值为1，减小到0.8以减少行距
\begin{table}[t]
\centering
\begin{tabular}{l c c c c c}
\toprule
\textbf{order} & \textbf{p=1} & \textbf{p=2} & \textbf{p=3} & \textbf{p=4} & \textbf{p=5} \\ \midrule
Cora & 88.96 & \textcolor{purple}{89.96} & 89.75 & 89.45 & N/A\\
Citeseer & 78.90 & 79.89 & \textcolor{purple}{79.93} & 79.84 & 79.84 \\
Actor & 39.90 & \textcolor{purple}{40.90} & 40.56 & 40.57 & 40.49 \\ 
Cornell & 87.40  & \textcolor{purple}{88.30} & 88.05 & N/A & N/A \\ 
\bottomrule
\end{tabular}
\caption{Node classification results of different orders}
\label{tab:order}
\end{table}

\paragraph{Analysis of Order $p$.}
The order $p$ represents the order in higher-order networks SCs. Table \ref{tab:order}) shows the results of order $p$ on node classification across Cora, Citeseer, Actor and Cornell four datasets. As observed in the table and also supported by findings in the previous literature~\cite{huang2024higher}, second order generally yields better performance. Hence, we consistently adopt second-order information in all our experiments to ensure fair and consistent evaluation.

\paragraph{Analysis of Teleportation Parameter $\alpha$.}
The teleportation parameter $\alpha$ represents the restart probability in HiPPR. All lines show relatively stable accuracy across the tested range, with minor fluctuations. The results on Wisconsin and Cora exhibit slight peaks near $\alpha=0.15$, while Citeseer remains consistently flat. In reference to common practice in most papers, we selected $\alpha=0.15$ as the optimal value.
\begin{figure}[t]
 \centering
  \includegraphics[width=3.2 in]{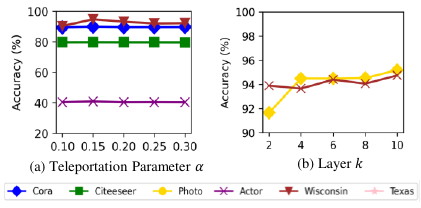}
    \caption{Visualization of sensitivity analysis.}
    \label{fig.sensitive}
\end{figure}

\begin{figure}[t]
 \centering
 \centering
  \includegraphics[width=3.2 in]{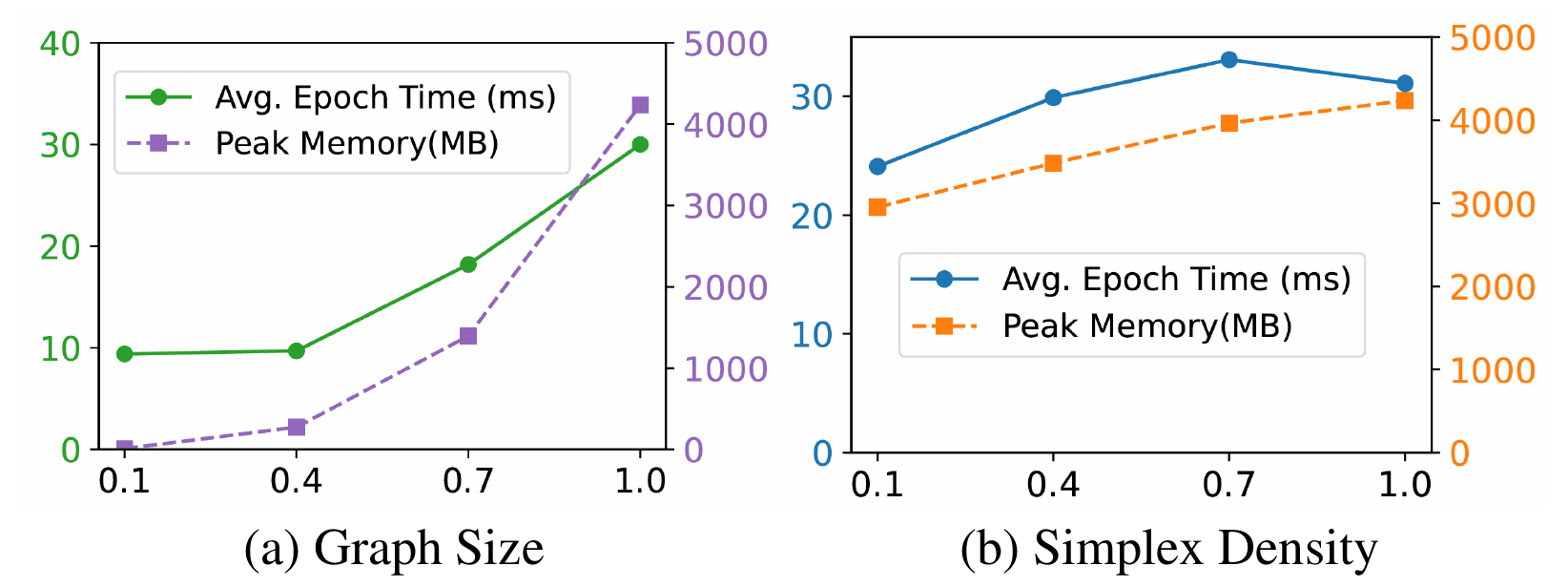}
    \caption{Training Time vs. Peak Memory Usage.}
    \label{fig.time_nd_memory}
\end{figure}
\begin{table}[b]
\centering
\setlength{\tabcolsep}{2pt}    % 调整列间距以适应新的布局
{\small % 开始调整字体大小
\begin{tabular}{l| c c c c c}
\toprule
Dataset & HPGNN & SNN & HIGCN & BernNet & GPRGNN \\ \midrule
Cora & 13.7/4.92 & 32.2/10.03 & 14.4/3.13 & 24.4/8.25 & 10.1/3.16 \\
Citeseer & 9.6/3.5 & 27.7/9.27 & 12.2/2.75 & 25.6/8.68 & 9.3/3.23 \\
Cornell & 8.6/2.02 & 28.1/7.33 & 10/2.18 & 23.7/7.52 & 9.7/2.61 \\
Wisconsin & 9.3/2.1 & 24.4/6.65 & 8.7/2 & 22.5/7.32 & 10.3/2.73 \\ 
\bottomrule
\end{tabular}
} % 结束调整字体大小
\caption{Efficiency on node classification experiments: Average running time per epoch(ms)/ average total running time(s).}
\label{tab:time}
\end{table}
% Explaination of the table

\paragraph{Analysis of Layer $k$.} 
The layer $k$ represents the layer in higher-order adaptive spectral graph convolutions. We investigated the impact of $k$ in HPGNN, evaluating its influence on model performance across Photo and Actor datasets. As shows in Table \ref{tab:order}(b), both lines show a general upward trend as $k$ increases, with minor fluctuations. we selected $k=10$ as the optimal value, as it yields the highest accuracy while maintaining stability across the tested range.

\subsection{Computationally Efficient}

Table~\ref{tab:time} illustrates that our model significantly outperforms Hodge-based SNN in terms of speed while maintaining performance comparable to pairwise methods. The runtime and memory usage curves for the Actor dataset, shown in Figure \ref{fig.time_nd_memory} highlight distinct trends. Specifically, it demonstrates that as the graph size increases, both the training time per epoch and peak memory usage exhibit exponential growth. In contrast, with an increase in simplex density, the growth in training time per epoch and peak memory usage follows a more linear trend. 

\subsection{Visualization}
\begin{figure}[t]
 \centering
  \includegraphics[width=3.2 in]{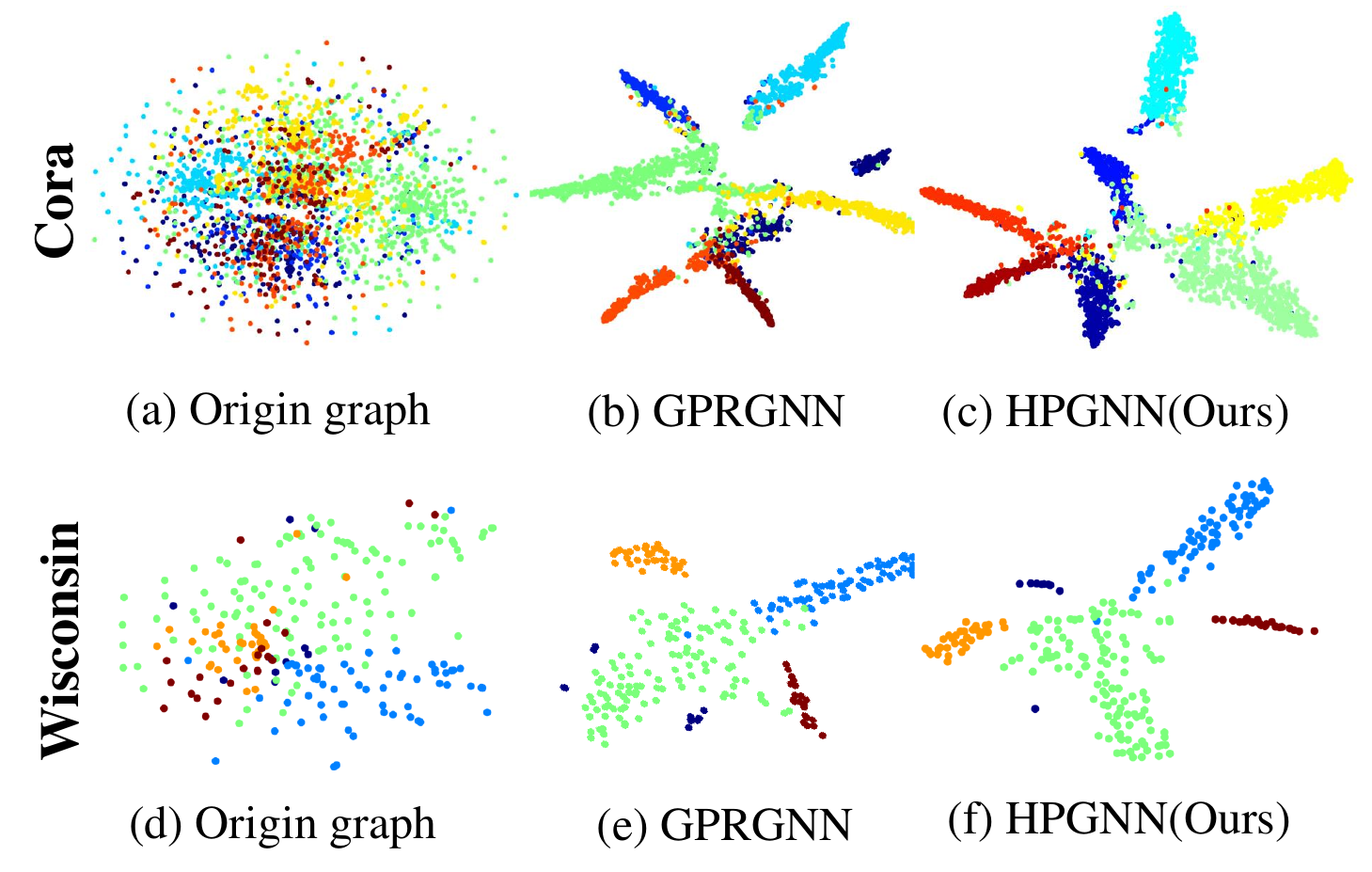}
    \caption{Structure visualization}
    \label{fig.tsne}
\end{figure}

To evaluate the clustering behavior and effectiveness of our proposed model, HPGNN, we present t-SNE visualizations for the Cora and Wisconsin datasets in Figure~\ref{fig.tsne}. These plots show how well the models separate classes in a low-dimensional space, with edges omitted for clarity.

As shows Figure~\ref{fig.tsne}(c), for the Cora dataset, HPGNN clusters nodes more clearly than GPRGNN, indicating that its integration of high-order information and PPR mechanisms improves the separation of classes in homophilic graphs. Figure~\ref{fig.tsne}(e) shows that in the Wisconsin dataset, HPGNN also outperforms other models in clustering heterophilic graphs, with better class separation, showing its robustness in handling complex structures.

\section{Conclusion}

In this paper, we propose HPGNN, a novel graph neural network model that combines higher-order node relationships with the personalized PageRank (PPR) mechanism to improve node classification, especially in complex graph structures. Unlike previous methods focusing on pairwise relationships, our approach captures multilevel interactions, crucial for heterophilic environments. 

Experiments on seven real-world datasets show that HPGNN outperforms state-of-the-art methods in 5 out of 7 cases, especially on heterophilic graphs. Its ability to handle noisy and complex relationships sets it apart, offering a more robust solution for graph representation learning. Further optimization is needed for simpler or sparse higher-order graphs to improve performance across all scenarios.

\section*{Acknowledgments}
This work was supported by the National Natural Science Foundation of China (No.62472305 and No.62272338), the Key R\&D Project in Hainan Province, China (No.ZDYF2024SHFZ051), and the Hainan Tropical Ocean Institute Yazhou Bay Innovation Research Institute
Major Science and Technology Plan Project (No.2023CXYZD001).

\section*{Contribution Statement}
Yumeng Wang and Zengyi Wo contributed equally to this work. Minglai Shao is the corresponding author.

\appendix

\clearpage % 再次开始新的一页

%% The file named.bst is a bibliography style file for BibTeX 0.99c
\bibliographystyle{named}
% \nocite{*} %% all references

\bibliography{ijcai25}

\end{document}